\documentclass{ws-rv9x6}
\usepackage{subfigure}
\usepackage{ws-rv-thm}
\usepackage{ws-rv-van}
\usepackage{ws-book-har}
\usepackage{overpic}
\usepackage{url}
\usepackage{fancyvrb}
\usepackage{amsmath}
\usepackage{amsmath, amssymb}
\usepackage{tikz}
\usepackage{algorithm}
\usepackage[noend]{algpseudocode}
\usepackage[multiple]{footmisc}
\usepackage{tikz}
\usetikzlibrary{shapes,arrows}
\makeatletter
\def\BState{\State\hskip-\ALG@thistlm}
\makeatother
\begin{document}
\chapter{Hybrid computer approach to train a machine learning system}
\renewcommand {\thefootnote}{\number\value{footnote}} 
\newcommand{\D}{\mathrm{d}}
\newcommand{\argmax}{\mathrm{argmax}}
 \author[M. Holzer, B. Ulmann]{\textsc{Mirko Holzer},\footnote{
  Mr. \textsc{Holzer} is co-founder and CEO of BrandMaker.} 
  \textsc{Bernd Ulmann}\footnote{Dr. \textsc{Ulmann} is professor for business
  informatics at the FOM University of Applied Sciences for Economics and 
  Management in Frankfurt/Main (Germany) and a guest professor and lecturer
  at the Institute of Medical Systems Biology at Ulm University.}}
 \begin{abstract}
  This chapter describes a novel approach to training machine learning systems
  by means of a hybrid computer setup i.\,e. a digital computer tightly coupled
  with an analog computer. In this example, a reinforcement learning system is
  trained to balance an inverted pendulum which is simulated on an analog 
  computer, demonstrating a solution to the major challenge of adequately 
  simulating the environment for reinforcement learning.\footnote{The
  analog/hybrid approach to this problem has also been described
  in \cite[sec.~6.24/7.4]{ulmann} with a focus on the analog computer
  part.}\footnote{The authors would like to thank Dr.~\textsc{Chris Giles} and
  Dr.~\textsc{David Farag\'{o}} for proof reading and making many invaluable
  suggestions and corrections which greatly enhanced this chapter.}
 \end{abstract}
 \section{Introduction}
  The following sections introduce some basic concepts which underpin the 
  remaining parts of this chapter.
  \subsection{A brief introduction to artificial intelligence
   and machine learning}  
   \emph{Machine learning} is one of the most exciting technologies of our
   time. It boosts the range of tasks that computers can perform to a level
   which has been extremely difficult, if not impossible, to achieve using
   conventional algorithms. 
   
   Thanks to machine learning, computers understand spoken language, offer
   their services as virtual assistants such as Siri or Alexa, diagnose cancer
   from magnetic resonance imaging (MRI), drive cars, compose music, paint
   artistic pictures, and became world champion in the board game Go.
   
   The latter feat is even more impressive when one considers that Go is
   probably one of the most complex game ever devised\footnote{We are
   referring here to games with complete information, whereas games with
   incomplete information, such as Stratego, are yet to be conquered by
   machine learning.}; for example it is much more complex than chess.
   Many observers in 1997 thought that IBM's \textsc{Deep Blue}, which
   defeated the then world chess champion \textsc{Garry Kasparov}, was
   the first proof-of-concept for Artificial Intelligence.
   However, since chess is a much simpler game than Go,
   chess-playing computer algorithms can use try-and-error methods
   to evaluate the best next move from the set of all possible sensible moves.
   In contrast, due to the $10^{360}$ possible game paths of Go, which is far
   more than the number of atoms in the universe, it is impossible
   to use such simple brute-force computer algorithms to calculate the best
   next move. This is why popular belief stated that it required human
   intuition, as well as creative and strategic thinking, to master Go --
   until Google's \textsc{AlphaGo} beat 18-time world champion
   \textsc{Lee Sedol} in 2016.
   
   How could \textsc{AlphaGo} beat \textsc{Lee Sedol}? Did Google develop
   human-like Artificial Intelligence with a masterly intuition for Go and the
   creative and strategic thinking of a world champion? Far from it.
   Google's \textsc{AlphaGo} relied heavily on a machine learning technique
   called \emph{reinforcement learning}, \emph{RL} for short, which is at the 
   center of the novel hybrid analog/digital computing approach introduced in 
   this chapter.
   
   The phrase \emph{Artificial Intelligence} (\emph{AI}) on the other hand,
   is nowadays heavily overused and frequently misunderstood. At the time of
   writing in early 2020, there exists no AI in the true sense of the word.
   Three flavors of AI are commonly identified:
   \begin{description}
    \item [Artificial Narrow Intelligence (ANI):] Focused on one narrow task
     such as playing Go, performing face recognition, or deciding the credit
     rating of a bank's customer. Several flavors of machine learning are used
     to implement ANI.
    \item [Artificial General Intelligence (AGI):] Computers with AGI would be
     equally intelligent to humans in every aspect and would be capable of
     performing the same kind of intellectual tasks that humans perform with
     the same level of success. Currently it is not clear if machine learning
     as we know it today, including \emph{Deep Learning}, can ever evolve into
     AGI or if new approaches are required for this.
   \item[Artificial Super Intelligence (ASI):] Sometimes called
    ``humanity's last invention'', ASI is often defined as an intellect that
    is superior to the best human brains in practically every field,
    including scientific creativity, general wisdom, and social skills. From
    today's perspective, ASI will stay in the realm of science fiction for
    many years to come.
   \end{description}
   Machine learning is at the core of all of today's AI/ANI efforts. Two
   fundamentally different categories of machine learning can be identified
   today: The first category, which consists of \emph{supervised learning} and
   \emph{unsupervised learning}, performs its tasks on existing data sets. As
   described in \cite[p.~85]{domingos12} in the context of Supervised Learning:
   \begin{quotation}
    ``\emph{A dumb algorithm with lots and lots of data beats a clever one with 
     modest amounts of it. (After all, machine learning is all about letting 
     data do the heavy lifting.)}''
   \end{quotation}
   \emph{Reinforcement learning} constitutes the second
   category. It is inspired by behaviorist psychology and does not rely on
   data. Instead, it utilizes the concept of a software \emph{agent} that can
   learn to perform certain \emph{actions}, which depend on a given
   \emph{state} of an \emph{environment}, in order to maximize some kind of
   long-term (cumulative) \emph{reward}.
   Reinforcement learning is very well suited for all kinds of control tasks,
   as it does not need sub-optimal actions to be explicitly corrected. The
   focus is on finding a balance between exploration (of uncharted territory)
   and exploitation (of current knowledge).

   Here are some typical applications of machine learning arranged by paradigm:
   \begin{description}
    \item [Supervised learning:] \emph{classification} (image classification,
     customer retention, diagnostics) and \emph{regression} (market
     forecasting, weather forecasting, advertising popularity prediction)
    \item [Unsupervised learning:] \emph{clustering} (recommender systems,
     customer segmentation, targeted marketing) and \emph{dimensionality
     reduction} (structure discovery, compression, feature elicitation)
    \item [Reinforcement learning:] robot navigation, real-time decisions,
     game AI, resource management, optimization problems
   \end{description}

   For the sake of completeness, \emph{(deep) neural networks} need to be
   mentioned, even though they are not used in this chapter. They are a very
   versatile family of algorithms that can be used to implement all three of
   supervised learning, unsupervised learning and reinforcement learning.
   When reinforcement learning is implemented using deep neural networks then
   the term \emph{deep reinforcement learning} is used. The general public
   often identifies AI or machine learning with neural networks; this is a
   gross simplification. Indeed there are plenty of other approaches for
   implementing the different paradigms of machine learning as described in
   \cite{domingos15}.
  \subsection{Analog vs. digital computing}
   Analog and digital computers are two fundamentally different approaches to 
   computation. A traditional digital computer, more precisely a 
   \emph{stored program digital computer}, has a fixed internal structure, 
   consisting of a control unit, arithmetic units etc., which are controlled 
   by an \emph{algorithm} stored as a \emph{program} in some kind of
   \emph{memory}. This algorithm is then executed in a basically stepwise 
   fashion. An analog computer, in contrast, does not have a memory at all and
   is not controlled by an algorithm. It consists of a multitude of 
   \emph{computing elements} capable of executing basic operations such as 
   addition, multiplication, integration (sic!) etc. These computing elements
   are then interconnected in a suitable way to form an \emph{analogue}, 
   a model, of the problem to be solved.

   So while a digital computer has a fixed internal structure and a variable
   program controlling its overall operation, an analog computer has a variable
   structure and no program in the traditional sense at all. The big advantage
   of the analog computing approach is that all computing elements involved in 
   an actual program are working in full parallelism with no data dependencies,
   memory bottlenecks etc. slowing down the overall operation. 

   Another difference is that values within an analog computer are typically
   represented as voltages or currents and thus are as continuous as possible
   in the real world.\footnote{There exist digital analog computers, which is 
   not the contradiction it might appear to be. They differ from classical
   analog computers mainly by their use of a binary value representation.
   Machines of this class are called \emph{DDA}s, short for \emph{Digital
   Differential Analyzer}s and are not covered here.} Apart from continuous 
   value representation, analog computers feature integration over 
   time-intervals as one of their basic functions.

   When it comes to the solution or simulation of systems described by 
   coupled differential equations, analog computers are much more energy 
   efficient and typically also much faster than digital computers. On the 
   other hand, the generation of arbitrary functions (e.\,g. functions, which 
   can not be obtained as the solution of some differential equation),
   complex logic
   expressions, etc. are hard to implement on an analog computer. So the idea
   of coupling a digital computer with an analog computer, yielding a 
   \emph{hybrid computer}, is pretty obvious. The analog computer basically 
   forms a high-performance co-processor for solving problems based on
   differential equations, while the digital computer supplies initial 
   conditions, coefficients etc. to the analog computer, reads back values and
   controls the overall operation of the hybrid system.\footnote{More details
   on analog and hybrid computer programming can be found in \cite{ulmann}.}
  \subsection{Balancing an inverse pendulum using reinforcement learning}
   Reinforcement learning is particularly well suited for an analog/digital
   hybrid approach because the RL \emph{agent} needs a simulated or real
   environment in which it can perform, and analog computers excel in
   simulating multitudes of scenarios. The inverse pendulum, as shown in
   figure \ref{pic_pendulum}, was chosen for the validation of the approach
   discussed in this chapter because it is one of the classical ``Hello World''
   examples of reinforcement learning and it can be easily simulated
   on small analog computers.

\begin{figure}
    \centering
    \includegraphics[width=.8\textwidth]{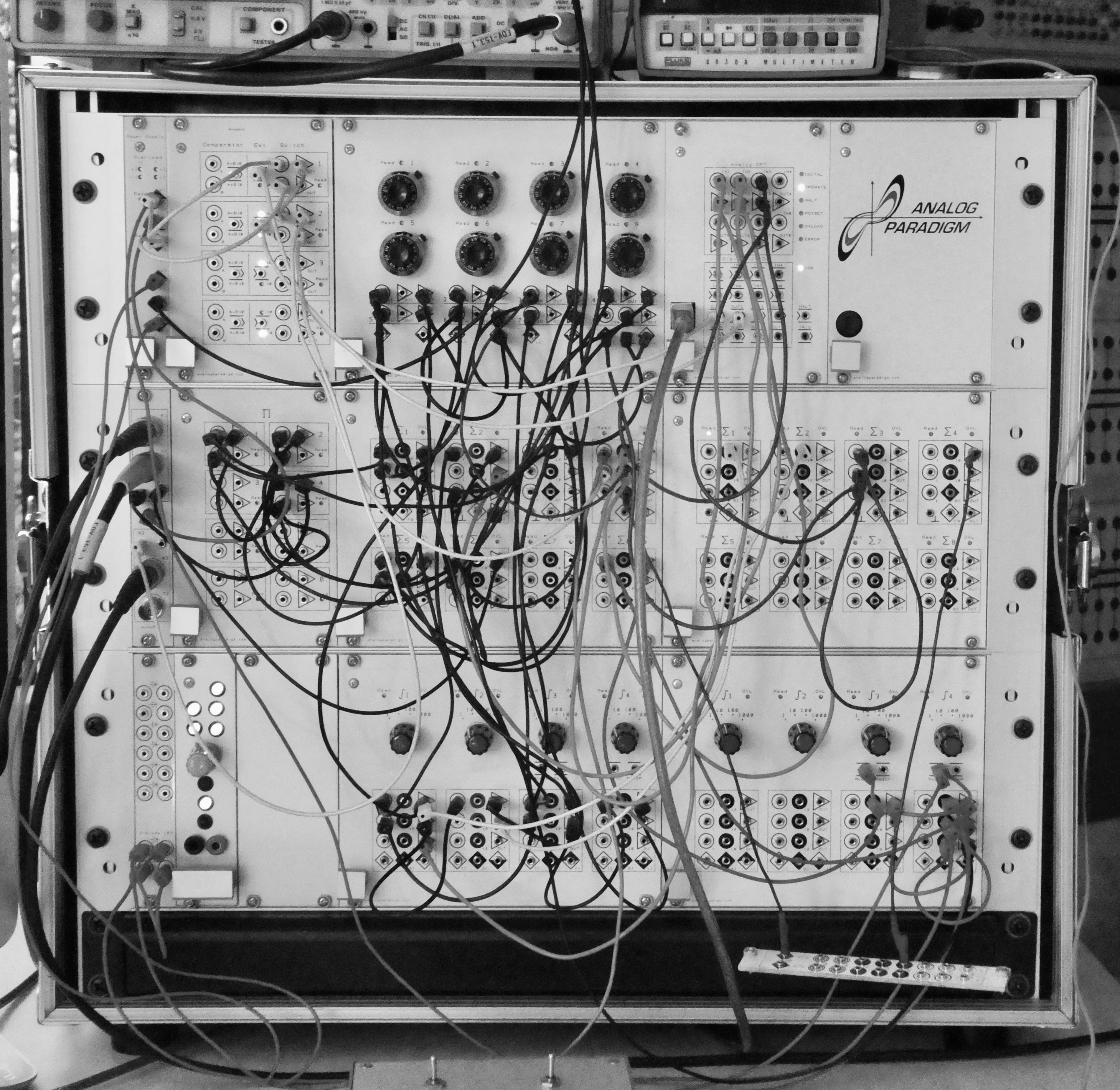}
    \caption{Analog computer setup}
    \label{pic_ac_setup}
   \end{figure}

   The hybrid computer setup consists of the analog computer shown in
   figure \ref{pic_ac_setup} that simulates the inverse pendulum and a digital
   computer running a reinforcement learning algorithm written in Python.

\begin{figure}
    \centering
    \vspace*{2mm}
    \scalebox{.8}{
    \tikzstyle{block} = [draw, fill=gray!10, rectangle,
     minimum height=8em, minimum width=6em]
    \tikzstyle{input} = [coordinate]
    \tikzstyle{output} = [coordinate]
    \tikzstyle{pinstyle} = [pin edge={to-,thin,black}]
    \begin{tikzpicture}[auto, node distance=2cm,>=latex']
     \node [block] (ac) {Analog computer};
     \node [block, right of=ac, node distance=6cm] (hc) {Hybrid controller};
     \node [block, right of=hc, node distance=5cm] (dc) {Digital computer};
     \draw [<-, transform canvas={yshift=3em}] (ac) -- node{\scriptsize mode control} (hc);
     \draw [->, transform canvas={yshift=1.5em}] (ac) -- node{\scriptsize analog readout} (hc);
     \draw [<-, transform canvas={yshift=0em}] (ac) -- node{\scriptsize digital outputs} (hc);
     \draw [->, transform canvas={yshift=-1.5em}] (ac) -- node{\scriptsize digital inputs} (hc);
     \draw [<-, transform canvas={yshift=-3em}] (ac) -- node{\scriptsize digital pot. ctrl.} (hc);
     \draw [<->, very thick] (hc) -- node{USB} (dc);
    \end{tikzpicture}
    }
    \vspace*{-5mm}
    \caption{Basic structure of the hybrid computer}
    \label{pic_hc_basic}
   \end{figure}
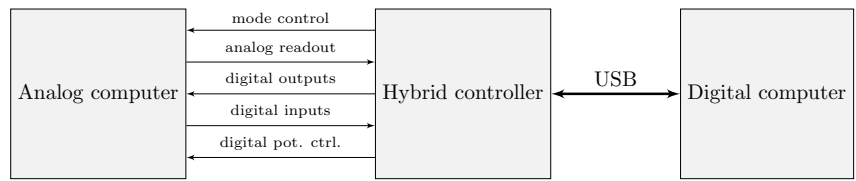

   Both systems communicate using serial communication over a USB connection.
   Figure \ref{pic_hc_basic} shows the overall setup: The link between the 
   digital computer on the right hand side and the analog computer is a 
   hybrid controller, which controls all parts of the analog computer, such 
   as the integrators, digital potentiometers for setting coefficients etc.
   This hybrid controller receives commands from the attached digital computer 
   and returns values read from selected computing elements of the analog 
   computer.
   The simulation and the learning algorithm both run in real-time\footnote{As 
   can be seen in this video: \protect\url{https://youtu.be/jDGLh8YWvNE}}.

   \label{def_episode} Reinforcement learning takes place
   in \emph{episodes}.  One episode is defined as ``balance the pendulum until
   it falls over or until the cart moves outside the boundaries of the
   environment''. The digital computer asks the analog computer for
   real-time simulation information such as the cart's $x$ position and the 
   pendulum's angle $\varphi$. The learning algorithm then decides if the 
   current episode, and therefore the current learning process, can continue or 
   if the episode needs to be ended; ending the episode also resets the
   simulation running on the analog computer.

 \section{The analog simulation} \label{s_analog}
\begin{figure}
   \centering
   \vspace*{-2cm}
   \includegraphics[width=.6\textwidth]{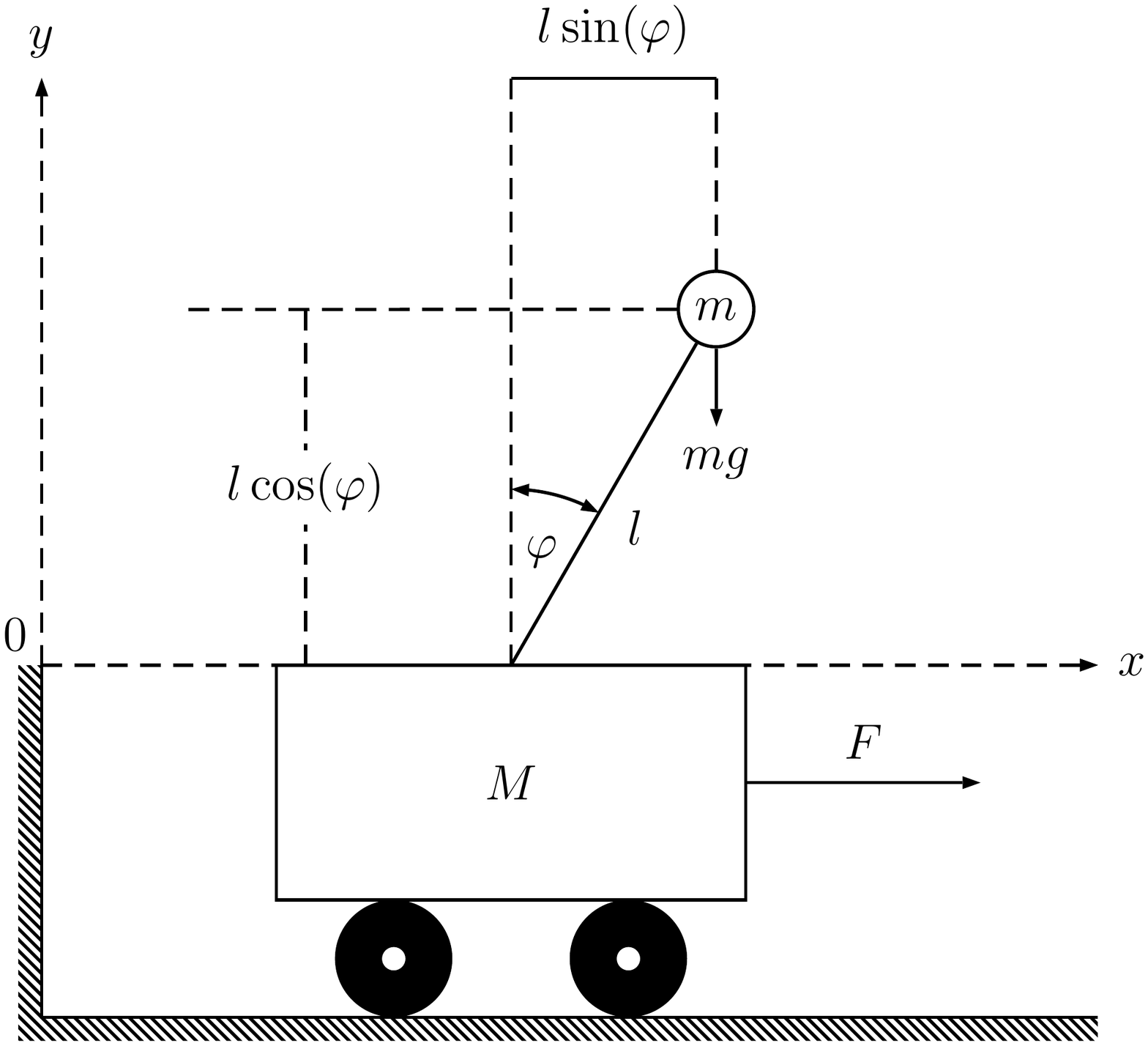}
   \vspace*{-2cm}
   \caption{Configuration of the inverted pendulum}
   \label{pic_pendulum}
  \end{figure}

  Simulating an inverted pendulum mounted on a cart with one degree of freedom,
  as shown in figure \ref{pic_pendulum}, on an analog computer is quite
  straightforward. The pendulum mass $m$ is assumed to be mounted on top of a 
  mass-less pole which in turn is mounted on a pivot on a cart which can be
  moved along the horizontal axis. The cart's movement is controlled by
  applying a force $F$ to the left or right side of the cart for a certain
  time-interval $\delta t$. If the cart couldn't move, the pendulum would
  resemble a simple mathematical pendulum described by 
  \[
   \ddot{\varphi}-\frac{g}{l}\sin(\varphi)=0
  \]
  where $\ddot{\varphi}$ is the second derivative of the pendulums angle
  $\varphi$ with respect to time.\footnote{In engineering notations like
  $\dot{\varphi}=\frac{\D\varphi}{\D t}$, 
  $\ddot{\varphi}=\frac{\D^2\varphi}{\D t^2}$ etc. are commonly used to denote 
  derivatives with respect to time.}

  Since in this example the cart is non-stationary, the problem is much more 
  complex and another approach has to be taken. In this case, the 
  \emph{Lagrangian}\footnote{This approach, which lies at the heart of 
  \emph{\textsc{Lagrangian} mechanics}, allows the use of so-called 
  \emph{generalized coordinates} like, in this case, the angle $\varphi$ and
  the cart's position $x$ instead of the coordinates in 
  classic \textsc{Newton}ian mechanics. More information on \textsc{Lagrangian} 
  mechanics can be found in \cite{brizard}.}
  \[
   L=T-V
  \]
  is used, where $T$ and $V$ represent the total kinetic and potential energy
  of the overall system. With $g$ representing the gravitational acceleration
  the potential energy is
  \[
   V=mgl\cos(\varphi),
  \]
  the height of the pendulum's mass above the cart's upper surface.

  The kinetic energy is the sum of the kinetic energies of the pendulum 
  bob with mass $m$ and the moving cart with mass $M$:
  \[
   T=\frac{1}{2}\left(Mv_{c}^2+mv_{p}^2\right),
  \]
  where $v_{\text{c}}$ and $v_{\text{p}}$ represent the velocities of the cart 
  and the pendulum, respectively. With $x$ denoting the horizontal position of 
  the cart, the cart's velocity $v_{\text{c}}$ is just the first derivative of 
  its position $x$ with respect to time:
  \[
   v_{\text{c}}=\dot{x}
  \]

  The pendulum's velocity is a bit more convoluted since the velocity of the 
  pendulum mass has two components, one along the $x$- and one along the 
  $y$-axis forming a two-component vector. The scalar velocity is then the 
  \textsc{Euclid}ean norm of this vector, i.\,e.
  \[
   v_{p}=\sqrt{\left(\frac{\text{d}}{\text{d}t}\left(x-l\sin(\varphi)\right)\right)^2+\left(\frac{\text{d}}{\text{d}t}\left(l\cos(\varphi)\right)\right)^2}.
  \]

  The two components under the square root are
  \[
   \left(\frac{\text{d}}{\text{d}t}\left(x-l\sin(\varphi)\right)\right)^2=
   \left(\dot{x}-l\dot{\varphi}\cos(\varphi)\right)^2=
    \dot{x}^2-2\dot{x}l\dot{\varphi}\cos(\varphi)+l^2\dot{\varphi}^2\cos^2(\varphi)
  \]
  and
  \[
   \left(\frac{\text{d}}{\text{d}t}\left(l\cos(\varphi)\right)\right)^2=
   \left(-l\dot{\varphi}\sin(\varphi)\right)^2=l^2\dot{\varphi}^2\sin^2(\varphi)
  \]
  resulting in
  \begin{align}
   v_{p}&=
   \dot{x}^2-2\dot{x}l\dot{\varphi}\cos(\varphi)+l^2\dot{\varphi}^2\cos^2(\varphi)+
   l^2\dot{\varphi}^2\sin^2(\varphi)\nonumber\\
   &=\dot{x}^2-2\dot{x}\dot{\varphi}l\cos(\varphi)+l^2\dot{\varphi}^2.\nonumber
  \end{align}

  This finally yields the Lagrangian
  \begin{align}
   L&=\frac{1}{2}M\dot{x}^2+\frac{1}{2}\left(\dot{x}^2-2\dot{x}\dot{\varphi}l\cos(\varphi)+l^2\dot{\varphi}^2\right)-mgl\cos{\varphi}\nonumber\\
    &=\frac{1}{2}(M+m)\dot{x}^2+m\dot{x}\dot{\varphi}l\cos(\varphi)+\frac{1}{2}ml^2\dot{\varphi}^2-mgl\cos(\varphi).\nonumber
  \end{align}

  As a next step, the \textsc{Euler}-\textsc{Lagrange}-equations
  \begin{align}
   \frac{\text{d}}{\text{d}t}\left(\frac{\partial L}{\partial\dot{x}}\right)-\frac{\partial L}{\partial x}&=F\text{~and}\label{equ_el_1}\\
   \frac{\text{d}}{\text{d}t}\left(\frac{\partial L}{\partial\dot{\varphi}}\right)-\frac{\partial L}{\partial\varphi}&=0\label{equ_el_2}
  \end{align}
  are applied. The first of these equations requires the following partial
  derivatives
  \begin{align*}
   \frac{\partial L}{\partial x}&=0\\
   \frac{\partial L}{\partial\dot{x}}&=(M+m)\dot{x}-ml\dot{\varphi}\cos(\varphi)\\
   \frac{\text{d}}{\text{d}t}\left(\frac{\partial L}{\partial\dot{x}}\right)&=(M+m)\ddot{x}-ml\ddot{\varphi}\cos(\varphi)+ml\dot{\varphi}^2\sin(\varphi)
  \end{align*}
  while the second equations relies on these partial derivatives:
  \begin{align*}
   \frac{\partial L}{\partial\varphi}&=ml\dot{x}\dot{\varphi}\sin(\varphi)+mgl\sin(\varphi)\\
   \frac{\partial L}{\partial\dot{\varphi}}&=-ml\dot{x}\cos(\varphi)+ml^2\dot{\varphi}\\
   \frac{\text{d}}{\text{d}t}\left(\frac{\partial L}{\partial\dot{\varphi}}\right)&=-ml\ddot{x}\cos(\varphi)+ml\dot{x}\dot{\varphi}\sin(\varphi)+ml^2\ddot{\varphi}
  \end{align*}

  Substituting these into equations \eqref{equ_el_1} and \eqref{equ_el_2} yields
  the following two \textsc{Euler}-\textsc{Lagrange}-equations:
  \begin{equation}
   \frac{\text{d}}{\text{d}t}\left(\frac{\partial L}{\partial\dot{x}}\right)=
   (M+m)\ddot{x}-ml\ddot{\varphi}\cos(\varphi)+ml\dot{\varphi}^2\sin(\varphi)=
   F\label{equ_ip_motion_1}
  \end{equation}
  and
  \[
   \frac{\text{d}}{\text{d}t}\left(\frac{\partial L}{\partial\dot{\varphi}}\right)=-ml\ddot{x}\cos(\varphi)+ml\dot{x}\dot{\varphi}\sin(\varphi)+ml^2\ddot{\varphi}-ml\dot{x}\dot{\varphi}\sin(\varphi)-mgl\sin(\varphi)=0.
  \]

  Dividing this last equation by $ml$ and solving for $\ddot{\varphi}$ results
  in 
  \[
   \ddot{\varphi}=\frac{1}{l}\left(\ddot{x}\cos(\varphi)+g\sin(\varphi)\right)
  \]
  which can be further simplified to 
  \begin{equation}
   \ddot{\varphi}=\ddot{x}\cos(\varphi)+g\sin(\varphi)\label{equ_ip_motion_2}
  \end{equation}
  by assuming the pendulum having fixed length $l=1$.

  The two final equations of motion \eqref{equ_ip_motion_1} and 
  \eqref{equ_ip_motion_2} now fully describe the behaviour of the inverted 
  pendulum mounted on its moving cart, to which an external force $F$ may be 
  applied in order to move the cart and therefore the pendulum due to its 
  inertia.

  To simplify things further, it can be reasonably assumed that the mass $m$
  of the pendulum bob is negligible compared with the cart's mass $M$,
  so that \eqref{equ_ip_motion_1} can be rewritten as
  \begin{equation}
   M\ddot{x}=F.\label{equ_ip_motion_1_1}
  \end{equation}

\begin{sidewaysfigure}
   \centering
   \begin{overpic}[angle=90,width=\textheight,unit=1mm,tics=10,]{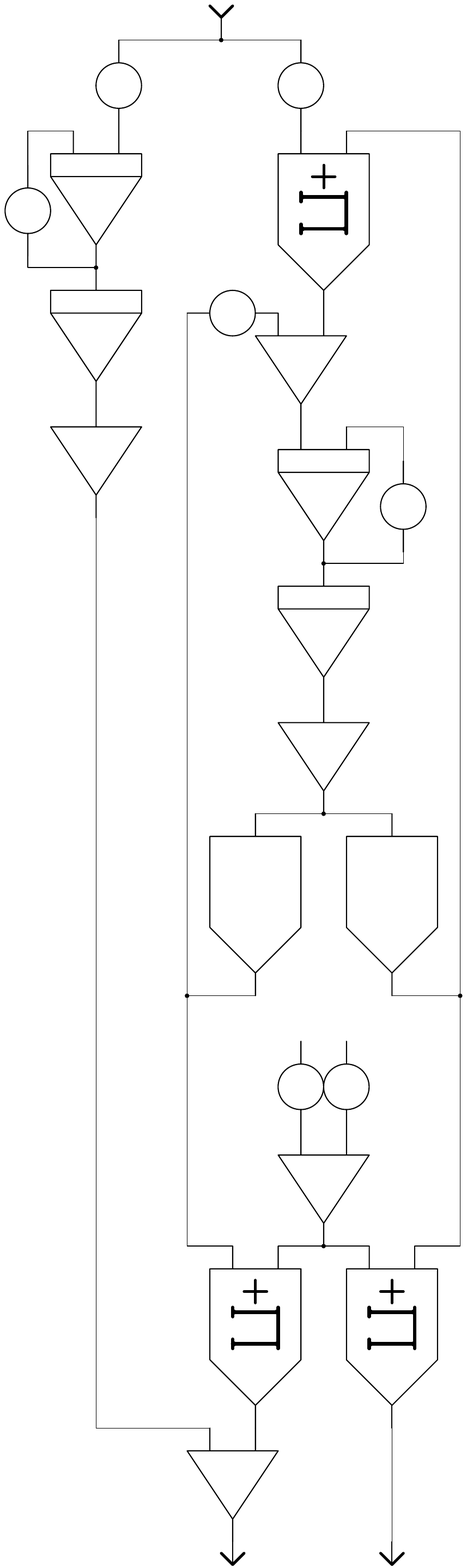}
    \put(.5, 35){$\ddot{x}$}
    \put(6, 40.2){$\gamma_1$}
    \put(6, 28.8){$\gamma_1$}
    \put(14, 23){$\beta_x$}
    \put(16, 28){$-\dot{x}$}
    \put(26, 28){$x$}
    \put(33, 28){$-x$}
    \put(11, 48){$\cos(\varphi)$}
    \put(20.8, 35.8){$g$}
    \put(32.4, 46.4){$\beta_\varphi$}
    \put(23, 34){$\sin(\varphi)$}
    \put(36.5, 40){$\dot{\varphi}$}
    \put(43.5, 42.5){$-\varphi$}
    \put(54.5, 45.7){$\cos(\dots)$}
    \put(54.5, 37.1){$\sin(\dots)$}
    \put(69.3, 39.8){$h$}
    \put(69.3, 42.7){$h$}
    \put(64.2, 39.8){$-1$}
    \put(61, 42.8){\small$\sin(\omega t)$}
    \put(99, 47.7){$y$}
    \put(99, 34){$x$}
   \end{overpic}
   \vspace*{-3cm}
   \caption{Modified setup for the inverted pendulum}
   \label{pic_acp_circuit}
  \end{sidewaysfigure}

  This simplification comes at a cost: the movement of the pendulum bob 
  no longer influences the cart, as it would have been the case with a 
  non-negligible pendulum mass $m$. Nevertheless, this is not a significant
  restriction and is justified by the the resulting simplification of the
  analog computer program shown in figure \ref{pic_acp_circuit}.

  As stated before, an analog computer program is basically an interconnection 
  scheme specifying how the various computing elements are to be connected 
  to each other. The schematic makes use of several standard symbols:
  \begin{itemize}
   \item Circles with inscribed values represent \emph{coefficients}. 
    Technically these are basically variable voltage dividers, so a coefficient 
    must always lie in the interval $[0,1]$. The input value 
    $\ddot{x}$ (the force applied to the cart in order to move it on the 
    $x$-axis) is applied to two such voltage dividers, each set to values
    $\gamma_1$ and $\gamma_2$. 
   \item Triangles denote \emph{summers}, yielding the sum of all of its inputs
    at its output. It should be noted that summers perform an implicit sign
    inversion, so feeding two values $-h$ and $h\sin(\omega t)$ to a summer as
    shown in the right half of the schematic, yields an output signal of
    $-(h\sin(\omega t)-h)=h(1-sin(\omega t))$.
   \item Symbols labelled with $+\Pi$ denote multipliers while those labelled
    with $\cos(\dots)$ and $\sin(\dots)$ respectively are function generators.
   \item Last but not least, there are integrators denoted by triangles with a 
    rectangle attached to one side. These computing elements yield the time
    integral over the sum of their respective input values. Just as with
    summers, integrators also perform an implicit sign-inversion.
  \end{itemize}

  Transforming the equations of motion \eqref{equ_ip_motion_2} and
  \eqref{equ_ip_motion_1_1} into an analog computer program is typically 
  done by means of the \emph{\textsc{Kelvin} feedback technique}.\footnote{More
  information on this can be found in classic text books on analog computing
  or in \cite{ulmann}.}

\begin{figure}
   \centering
   \vspace*{-4cm}
   \begin{overpic}[width=.7\textwidth,unit=1mm,tics=10,]{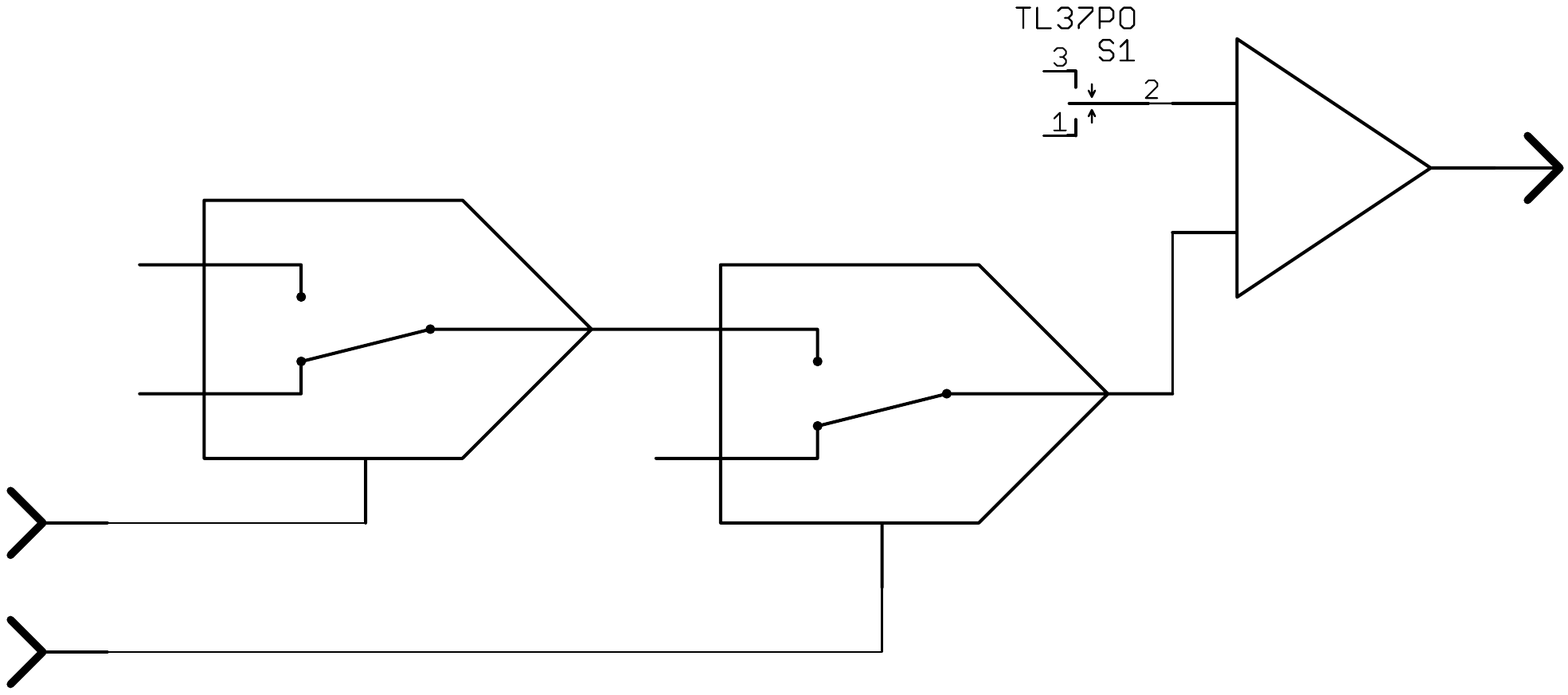}
    \put(16, 61.4){\scriptsize push the cart manually}
    \put(5.5, 48.8){$-1$}
    \put(5.5, 54.3){$+1$}
    \put(0, 37.7){\texttt{D0}}
    \put(0, 43.3){\texttt{D1}}
    \put(43, 60){$+1$}
    \put(43, 62.7){$-1$}
    \put(71, 58.5){$\ddot{x}$}
   \end{overpic}
   \vspace*{-4.5cm}
   \caption{Control circuit for the controlled inverted pendulum}
   \label{pic_acp_control}
  \end{figure}

  The tiny analog computer subprogram shown in figure \ref{pic_acp_control}
  shows how the force applied to the cart is generated in the hybrid computer
  setup. The circuit is controlled by two digital output signals from the hybrid
  controller, the interface between the analog computer and its digital 
  counterpart. Activating the output \texttt{D0} for a short time interval 
  $\delta t$ generates a force impulse resulting in an acceleration $\ddot{x}$
  of the cart. The output
  \texttt{D1} controls the direction of $\ddot{x}$, thus allowing the cart to 
  be pushed to the left or the right. Both of these control signals are 
  connected to electronic switches which are part of the analog computer.

  The outputs of these two switches are then fed to 
  a summer which also takes a second input signal from a manual operated 
  SPDT (single pole double throw) switch. This switch is normally open and 
  makes it possible for an operator to manually disturb the cart, thereby 
  unbalancing the pendulum. It is interesting to see how the RL system
  responds to such external disturbances.
 \section{The reinforcement learning system}
  Reinforcement learning, as defined in \cite[sec.~1.1]{sutton}, 
  \begin{quotation}
   ``\emph{is learning
   what to do --- how to map situations to actions --- so as to maximize a
   numerical reward signal. The learner (agent) is not told which actions to
   take, but instead must discover which actions yield the most reward by
   trying them. In the most interesting and challenging cases, actions may
   affect not only the immediate reward but also the next situation and,
   through that, all subsequent rewards. These two characteristics
   --- trial-and-error search and delayed reward --- are the two most important
   distinguishing features of reinforcement learning.}''
  \end{quotation}

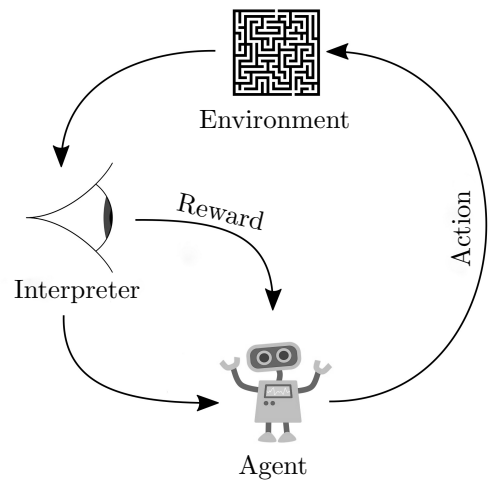
\begin{figure}
   \centering
   \begin{overpic}[width=.6\textwidth,unit=1mm,tics=10,]{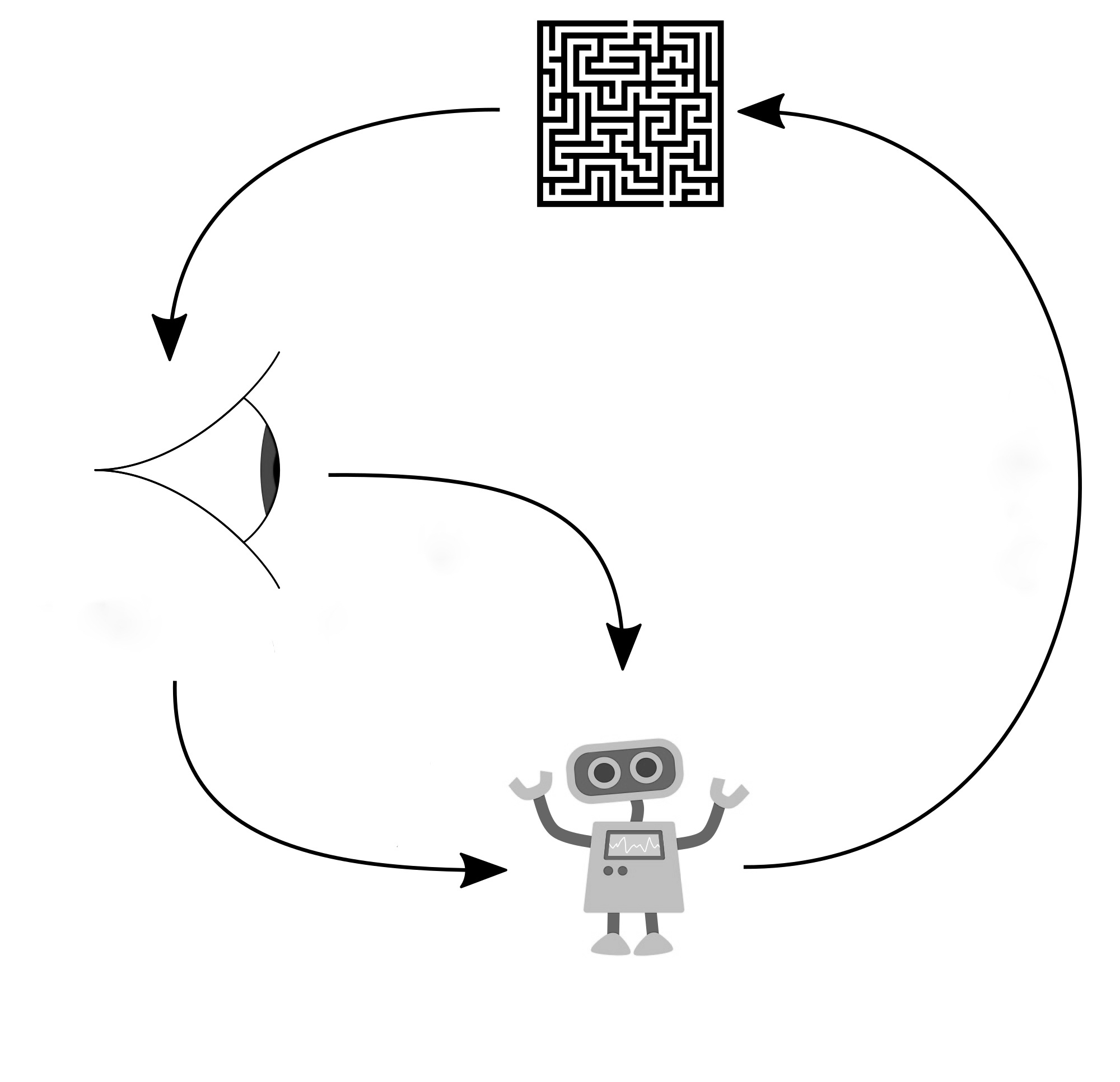}
    \put(41.5, 72){Environment}
    \put(37, 56){\rotatebox{-13}{Reward}}
    \put(6, 39){Interpreter}
    \put(49, 5){Agent}
    \put(90, 45){\rotatebox{90}{Action}}
   \end{overpic}
   \vspace*{-5mm}
   \caption{Reinforcement learning}
   \label{pic_rl_basics}
  \end{figure}

  In other words, reinforcement learning utilizes the concept of an
  \emph{agent} that can learn to perform certain \emph{actions} depending
  on a given \emph{state} of an \emph{enivronment} in order to maximize some
  kind of long-term (delayed) \emph{reward}. Figure 
  \ref{pic_rl_basics}\footnote{Source
  \protect\url{https://en.wikipedia.org/wiki/Reinforcement_learning},
  retrieved Jan. 9th, 2020.}
  illustrates  the interplay of the components of a RL system: In each
  \emph{episode}\footnote{See also section \ref{def_episode} for a definition
  of \emph{episode}.}, the agent observes the state $s \in \mathbb{S}$ of
  the system: position $x$ of the cart, speed $\dot{x}$, angle $\varphi$ of
  the pole and angular velocity $\dot{\varphi}$.
  Depending on the state, the agent performs
  an action that modifies the environment. The outcome of the action
  determines the new state and the short-term reward, which is the main
  ingredient in finding a \emph{value function} that is able to
  predict the long-term reward.

\begin{sidewaysfigure}
   \centering
   \hspace*{-2cm}
   \begin{tikzpicture}
    \node at(2,5.7) {Agent focuses on short term reward:};
 
    \draw[line width=1pt] (0,3.5) rectangle (2,4.5);
    \draw[line width=1pt] (.5,3.35) circle (3.2pt);
    \draw[line width=1pt] (1.5,3.35) circle (3.2pt);
    \draw[line width=1pt] (1,4.5) -- (0.293,5.207);
 
    \node at (4,4) {$\xrightarrow[r_\text{short}(s_i,a_i)=\text{max}]{\text{large $F<0$}}$};
 
    \draw[line width=1pt] (6,3.5) rectangle (8,4.5);
    \draw[line width=1pt] (6.5,3.35) circle (3.2pt);
    \draw[line width=1pt] (7.5,3.35) circle (3.2pt);
    \draw[line width=1pt] (7,4.5) -- (7,5.5);
 
    \node at (10,4) {$\xrightarrow[r_\text{short}(s_{i+1},a_{i+1})=\text{max}]{F=0}$};
 
    \draw[line width=1pt] (12,3.5) rectangle (14,4.5);
    \draw[line width=1pt] (12.5,3.35) circle (3.2pt);
    \draw[line width=1pt] (13.5,3.35) circle (3.2pt);
    \draw[line width=1pt] (13,4.5) -- (13.707,5.207);
 
    \node at (15,4) {\dots};
 
    \draw[line width=1pt] (16,3.5) rectangle (18,4.5);
    \draw[line width=1pt] (16.5,3.35) circle (3.2pt);
    \draw[line width=1pt] (17.5,3.35) circle (3.2pt);
    \draw[line width=1pt, dashed] (17,4.5) -- (16.293,5.207);
    \draw[line width=1pt, dashed] (17,4.5) -- (17.707,5.207);
 
    \node at (17,3) {\scriptsize The pendulum bounces};
    \node at (17,2.75) {\scriptsize back and forth with ever};
    \node at (17,2.45) {\scriptsize increasing angle};
    \node at(2,2.2) {Agent focuses on long term reward:};
 
    \draw[line width=1pt] (0,0) rectangle (2,1);
    \draw[line width=1pt] (.5,-.15) circle (3.2pt);
    \draw[line width=1pt] (1.5,-.15) circle (3.2pt);
    \draw[line width=1pt] (1,1) -- (0.293,1.707);
 
    \node at (4,.5) {$\xrightarrow[r_\text{short}(s_i,a_i)<\text{max}]{\text{small $F<0$}}$};
 
    \draw[line width=1pt] (6,0) rectangle (8,1);
    \draw[line width=1pt] (6.5,-.15) circle (3.2pt);
    \draw[line width=1pt] (7.5,-.15) circle (3.2pt);
    \draw[line width=1pt] (7,1) -- (6.6173,1.924);
 
    \node at (10,.5) {$\xrightarrow[r_\text{short}(s_{i+1},a_{i+1})<\text{max}]{\text{even smaller $F<0$}}$};
 
    \draw[line width=1pt] (12,0) rectangle (14,1);
    \draw[line width=1pt] (12.5,-.15) circle (3.2pt);
    \draw[line width=1pt] (13.5,-.15) circle (3.2pt);
    \draw[line width=1pt] (13,1) -- (13,2);
 
    \node at (15,.5) {\dots};
 
    \draw[line width=1pt] (16,0) rectangle (18,1);
    \draw[line width=1pt] (16.5,-.15) circle (3.2pt);
    \draw[line width=1pt] (17.5,-.15) circle (3.2pt);
    \draw[line width=1pt, dashed] (17,1) -- (16.6173,1.924);
    \draw[line width=1pt, dashed] (17,1) -- (17.382,1.924);
 
    \node at (17,-.5) {\scriptsize The pendulum bounces};
    \node at (17,-.75) {\scriptsize back and forth with ever};
    \node at (17,-1.05) {\scriptsize decreasing angle};
   \end{tikzpicture}  
   \caption{Rewards in the use-case of the inverted pendulum}
   \label{pic_rl_tikz}
  \end{sidewaysfigure}
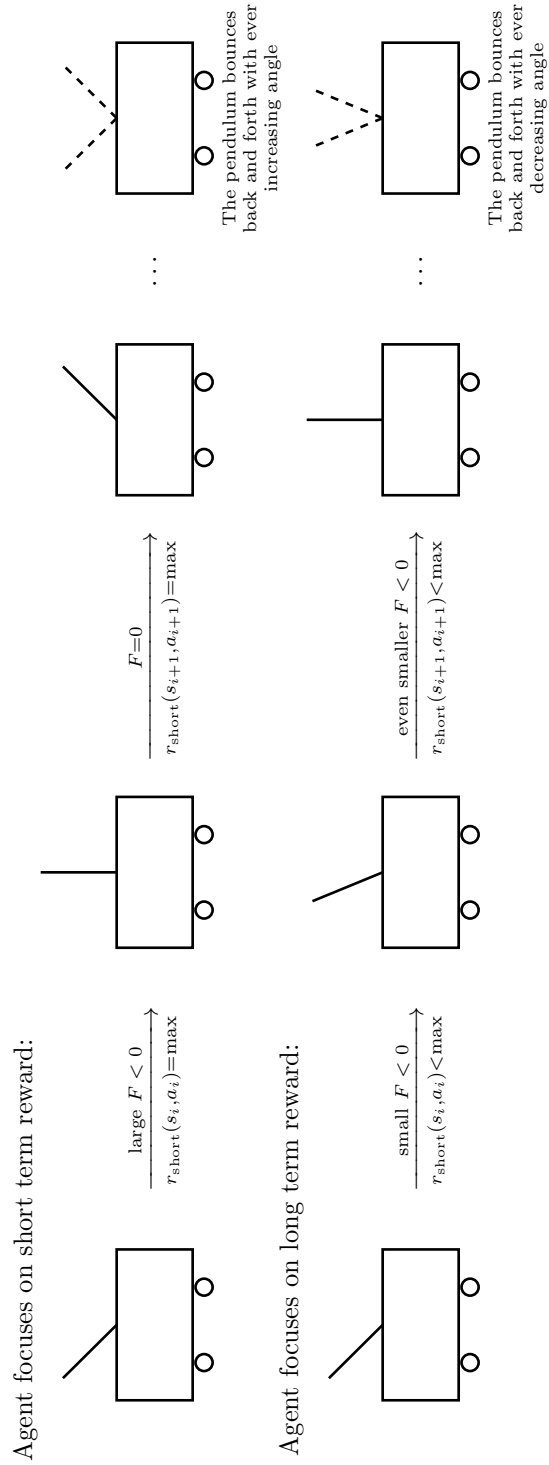

  Figure \ref{pic_rl_tikz} translates these abstract concepts into the
  concrete use-case of the inverted pendulum. The short-term reward is just
  a means to an end for finding the value function: Focusing on the short-term
  reward would only lead to an ever-increasing bouncing of the pendulum, equal
  to the control algorithm diverging. Instead, using the short-term reward to
  approximate the value function that can predict the long-term reward leads
  to a robust control algorithm (convergence).

  \subsection{Value function} \label{ss_valfunc}
   Roughly speaking, the \emph{state value function} $V$ estimates
   ``how benficial'' it is to be in a given state. The \emph{action value 
   function} $Q$ specifies ``how beneficial'' it is to take a certain action 
   while being in 
   a given state. When it comes for an agent to choose the next best action
   $a$, one straightforward \emph{policy} $\pi$ is to evaluate the action 
   value function $Q(s, a)$ for all actions $a$ that are possible while being
   in state $s$ and to choose the action $a_{\text{best}}$ yielding the maximum 
   value of the action value function:
   \[
    \pi \gets \text{always take}~\argmax_{a}(Q(s, a))~\text{as next action}
   \]
   A policy like this is also called a \emph{greedy policy}. 

   The value function $Q$ for a certain action $a$ given state $s$ is defined
   as the expected future reward $R$\footnote{The future reward $R$ in this
   chapter equates the long-term reward in the previous chapter.} that can be
   obtained when the agent is following the current policy $\pi$.
   The future reward is the sum of the
   discounted rewards $r$ for each step $t$ beginning from ``now''
   ($s_{0}=s$ at $t=0$) and ranging to the terminal state\footnote{The
   reward in the terminal state is defined to be zero, so that the infinite
   sum over bounded rewards always yields a finite value.}.
   $\gamma \in [0, 1]$ is implemented as a discount factor by exponentiating
   it by the number of elapsed steps $t$ in the current training episode.
   \[
    Q(s, a)=E[R(s, a)]=E\Bigg[\sum_{t=0}^{\infty}\gamma^{t}r_{t}\Big|_{s_{0}=s, a_{0}=a}\Bigg]
   \]

   Two important observations regarding the above-mentioned definition of $Q$
   are:
   \begin{enumerate}
    \item In step $t=0$ given the state $s_{0}$, the reward $r_{0}$ only
          depends on the action $a_{0}$. But all subsequent rewards
          $r_{t}\big|_{t>0}$ are depending on all actions that have been
          previously taken.
    \item $\pi$ and $Q(s,a)$ are not defined recursively, but iteratively.
   \end{enumerate}

   Following the policy $\pi$ as defined here is trivial if $Q$ is given since
   all that needs to be done in each step is to calculate
   $\argmax_{a}(Q(s, a))$.
   So the main challenge of reinforcement learning is to calculate the state
   value function $V_{\pi}(s)$ and then derive the action value function $Q$
   from it or to determine $Q$ directly. But often, $V$ or $Q$ cannot be
   calculated directly due to CPU or memory constraints. For example the state
   space
   $\mathbb{S} = \{(x, \dot{x}, \varphi, \dot{\varphi}) \in \mathbb{R}^4\}$
   used here is infinitely large. And even if it is discretized, the resulting
   combinatorial explosion still results in an enormous state space. So it
   cannot be expected to find an optimal value function in the limit of finite
   CPU and memory resources as shown in \cite[Part II]{sutton}. Instead, the
   goal is to find a good approximate solution that can be calculated
   relatively quickly.
  \subsection{$Q$-learning algorithm}
   There are several approaches to find the (action) value function $Q$. The
   implementation described here uses the \emph{$Q$-learning algorithm}, which
   is a model-free reinforcement learning algorithm. ``Model-free'' means, that
   it actually ``does not matter for the algorithm'', what the semantics of 
   the parameters/features of the environment are. It ``does not know''
   the meaning of a feature such as the cart position $x$ or the pole's
   angular velocity $\dot{\varphi}$. For the $Q$-learning algorithm, the set of 
   the features $(x, \dot{x}, \varphi, \dot{\varphi})$ is just what the
   current state within the environment comprises. And the state enables
   the agent to decide which action to perform next.

   Chris Watkins introduced $Q$-learning in \cite{watkins89} and
   in \cite[ch.~6]{sutton}. $Q$-learning is defined as a variant of
   \emph{temporal-difference learning} (TD): 
   \begin{quotation}
    ``\emph{TD learning is a combination of
     Monte Carlo ideas and dynamic programming (DP) ideas. Like Monte Carlo
     methods, TD methods can learn directly from raw experience without a model
     of the environment’s dynamics. Like DP, TD methods update estimates based
     in part on other learned estimates, without waiting for a final outcome
     (they bootstrap).}''
   \end{quotation}

   One of the most important properties of $Q$-learning is the concept of
   \emph{explore~vs.~exploit}: The learning algorithm alternates between
   \emph{exploiting} what it already knows and \emph{exploring} unknown
   territory. Before taking a new action, $Q$-learning evaluates a probability
   $\varepsilon$ which decides, if the next step is a random move (explore)
   or if the next move is chosen according to the policy $\pi$ (exploit). As
   described in section \ref{ss_valfunc}, in the context of this chapter the
   policy $\pi$ is a greedy policy that always chooses the best possible
   next action, i.e.\ $\argmax_{a}(Q(s, a))$.

   \begin{algorithm}
    \caption{$Q$-learning}\label{a_q-learning}
    \begin{algorithmic}[1]
     \Procedure{Find\_Action\_Value\_Function\_Q}{}
      \State $\alpha \gets \text{step size aka learning rate} \in (0, 1]$
      \State $\gamma \gets \text{discount rate} \in (0, 1]$
      \State $\varepsilon \gets \text{probability to explore,
                                   small and \textgreater~0}$
      \State $\text{Initialize}~Q(s, a)~\forall~s \in \mathbb{S},
              a \in \mathbb{A}(s), \text{arbitrarily}$
      \State $\text{~~~~except that}~Q(terminal~state, \cdot) = 0$
      \BState Loop for each episode:
      \State Initialize $s$
      \While{$s \neq terminal~state$} \label{a_qwhile}
       \If{$random~number < \varepsilon$}
        \State $a \gets \text{random action} \in \mathbb{A}(s)$
        \Comment{means: Explore}
       \Else
        \State $a \gets \argmax_{a}(Q(s, a))$
        \Comment{means: Exploit}
       \EndIf
       \State Take action $a$
       \State $r \gets \text{reward for taking action}~a~
               \text{while in state}~s$
       \State $s' \gets \text{state that follows due to taking action}~a$
       \State $a_{\text{best}} \gets \argmax_{a}(Q(s', a))$
       \State \label{a_qmagic} $Q(s, a) \gets 
               Q(s, a) + \alpha[r + \gamma * Q(s', a_{\text{best}}) - Q(s, a)]$
       \State $s \gets s'$
      \EndWhile
     \EndProcedure
    \end{algorithmic}
   \end{algorithm}

   Algorithm \ref{a_q-learning} is a temporal difference learner based on
   Bellman's equation and is inspired by \cite[sec.~6.5]{sutton}.
   The state space of the inverse pendulum is denoted as
   $\mathbb{S} = \{(x, \dot{x}, \varphi, \dot{\varphi}) \in \mathbb{R}^4\}$
   and $\mathbb{A}: S \rightarrow 2^A$ describes all
   valid\footnote{It is possible that not all actions that are theoretically
   possible in an environment are valid in all given states.} actions
   for all valid states.

   In plain english, the algorithm can be described as follows. For each step
   within an episode: Decide, if the next action shall explore new territory
   or exploit knowledge, that has already been learned. After this decision:
   Take the appropriate action $a$, which means that the reward $r$ for this
   action is collected and the system enters a new state $s'$ (see figure
   \ref{pic_rl_basics}). After that: Update the action value function
   $Q(s, a)$ with an \emph{estimated future reward} (line \ref{a_qmagic} of
   algorithm \ref{a_q-learning}), but discount the estimated future reward
   with the \emph{learning rate} $\alpha$.

   In machine learning in general the learning rate is a very
   important parameter, because it makes sure, that new knowledge that has
   been obtained in the current episode does not completely
   ``overwrite'' past knowledge.

   The calculation of the estimated future reward is obviously where the
   magic of $Q$-learning is happening:
   \[
    Q(s, a) \gets Q(s, a) + \alpha[r + \gamma * Q(s', a_{\text{best}}) - Q(s,a)]
   \]
   $Q$-learning estimates the future reward by taking the short-term reward $r$
   for the recently taken action $a$ and adding the discounted delta between
   the best possible outcome of a hypothetical next action $a_{\text{best}}$ 
   (given
   the new state $s'$) and the old status quo $Q(s, a)$. It is worth
   mentioning that adding the best possible outcome of $s'$ aka
   $a_{\text{best}} = \argmax_{a}(Q(s', a))$ can only be a guess,
   because it is unclear at the time of this calculation if action 
   $a_{\text{best}}$ is ever being taken due to
   the ``explore vs. exploit'' strategy.
   Still, it seems logical that if $Q(s, a)$ claims to be a measure of
   ``how beneficial'' a certain action is, that
   ``the best possible future'' $a_{\text{best}}$ from this starting point onwards
   is then taken into consideration when estimating the future reward.

   The discount rate $\gamma$ takes into consideration that the future is
   not predictable and therefore future rewards cannot be fully counted on.
   In other words, $\gamma$ is a parameter that balances the RL agent between
   the two opposite poles ``very greedy and short-sighted'' and
   ``decision making based on extremely long-term considerations.'' As such,
   $\gamma$ is one of many so called \emph{hyper parameters}; see also section
   \ref{sss_ftrans} et seq.

   Given the size of $\mathbb{S}$ it becomes clear that many episodes
   are needed to train a system using $Q$-learning. A rigorous proof that
   $Q$-learning works and that the algorithm converges is given in
   \cite{watkins92}.

   When implementing $Q$-learning, a major challenge is how to represent the
   function $Q(s, a)$. For small to medium sized problems that can be
   discretized with reasonable effort, tabular representations are
   appropriate. But as soon as the state space $\mathbb{S}$ is large, as
   in the case of the inverse pendulum, other representations
   need to be found.
  \subsection{Python implementation}
   The ideas of this chapter culminate in a hybrid analog/digital machine
   learning implementation that uses reinforcement learning, more specifically
   the $Q$-learning algorithm, in conjunction with linear regression, to solve
   the challenge of balancing the inverse pendulum.

   This somewhat arbitrary choice by the authors is not meant to claim
   that using RL and $Q$-learning is the only or at least optimal way of
   balancing the inverse pendulum. On the contrary, it is actually more
   circuitous than many other approaches. Experiments performed by the authors
   in Python have shown that a simple random search over a few thousand tries
   yields such $\vec{\theta} = (\theta_1, \theta_2, \theta_3, \theta_4)$, 
   that the result of the dot-multiplication with the current state
   of the simulation $\vec{s} = (x, \dot{x}, \varphi, \dot{\varphi})$ can
   reliably decide if the cart should be pushed to the left or to the right,
   depending, on whether or not
   $\label{ss_f_theta_s}\vec{\theta} \cdot \vec{s}$ is larger than zero.

   So the intention of this chapter is something else. The authors wanted to
   show that a general purpose machine learning algorithm like $Q$-learning,
   implemented as described in this chapter, is suitable without change
   for a plethora of other (and much more complex) real world applications and
   that it can be efficiently trained using a hybrid analog/digital setup.

   Python offers a vibrant Open Source ecosystem of building-blocks for
   machine learning, from which \textsc{scikit-learn}, as described in
   \cite{pedregosa}, was chosen as the foundation of our
   implementation\footnote{\label{fn_github}Full source code on GitHub:
   \protect\url{https://git.io/Jve3j}}.
   \subsubsection{States} \label{sss_states}
    Each repetition of the \texttt{While} loop in line \ref{a_qwhile} of
    algorithm \ref{a_q-learning} represents one step within the current
    iteration. The \texttt{While} loop runs until the current state
    $s \in \mathbb{S}$ of the inverse pendulum's simulation reaches a
    \emph{terminal state}. This is when the simulation of one episode ends,
    the simulation is reset and the next episode begins. The terminal state is
    defined as the cart reaching  a certain position $\left|x\right|$ which
    can be interpreted as ``the cart moved too far and leaves the allowed
    range of movements to the left or to the right or bumps into a wall''.
    Another trigger for the terminal state is that the angle
    $\left|\varphi\right|$ of the pole is greater than a certain predefined
    value, which can be interpreted as ``the pole falls over''.

    In Python, the state $s \in \mathbb{S}$ is represented as a standard
    Python \texttt{Tuple} of \texttt{Floats} containing the four elements
    $(x, \dot{x}, \varphi, \dot{\varphi})$ that constitute the state.

    During each step of each episode, the analog computer is queried for the
    current state using the \texttt{hc\_get\_sim\_state()} function (see also
    sec. \ref{ss_hybrid}). On the analog computer, the state is by the nature
    of analog computers a continuous function. As described in
    \cite[pp. 3--42]{wiering} an environment in reinforcement learning is
    typically stated in the form of a Markov decision process (MDP), because
    $Q$-learning and many other reinforcement learning algorithms
    utilize dynamic programming techniques. As an MDP is a discrete
    time stochastic control process, this means that for the Python
    implementation, we need to discretize the continuous state representation
    of the analog computer. Conveniently, this is happening automatically,
    because the repeated querying inside the \texttt{While} loop mentioned
    above is nothing else than \emph{sampling} and therefore a discretization
    of the continuous (analog) state.\footnote{Due to the partially
    non-deterministic nature of the Python code execution, the sampling rate
    is not guaranteed to be constant. The slight jitter introduced by this
    phenomenon did not notably impede the $Q$-learning.}
   \subsubsection{Actions} \label{ss_actions}
    In theory, one can think of an infinite\footnote{Example: Push the cart
    from the left with force 1, force 2, force 3, force 4, \dots} number of
    actions that can be performed on the cart on which the inverted pendulum
    is mounted. As this would complicate the implementation of the $Q$-learning
    algorithm, the following two simplifications where chosen while
    implementing actions in Python:
    \begin{itemize}
     \item There are only two possible  actions $a \in \{0, 1\}$:
      ``Push the cart to the left'' and ``push the cart to the right''.
     \item Each action $a$ is allowed in all states $s \in \mathbb{S}$
    \end{itemize}
    The actions $a=0$ and $a=1$ are translated to the analog computer
    simulation by applying a constant force from the right (to push the cart
    to the left) or the other way round for a defined constant period of time.
    The magnitude of the force that is being applied is configured as a
    constant input for the analog computer using a potentiometer, as described
    in section \ref{s_analog}. The constant period of time for which the force
    is being applied can be configured in the Python software module
    using \texttt{HC\_IMPULSE\_DURATION}.
    \subsubsection{Modeling the action value function $Q(s, a)$} \label{sss_Q}
     A straightforward way of modeling the action value function in Python
     could be to store $Q(s, a)$ in a Python \texttt{Dictionary}, so that the
     Python equivalent of line \ref{a_qmagic} in algorithm \ref{a_q-learning}
     would look like this:
     \begin{Verbatim}[numbers=left,frame=single,labelposition=all,
                      label=\scriptsize dictionary.py]
Q_s_a = {} #create a dictionary to represent Q(s, a)

[...]      #perform the Q-learning algorithm

#learn by updating Q(s, a) using learning rate alpha
Q_s_a[((x, xdot, phi, phidot), a)] = old_q_s_a + 
                                     alpha * predicted_reward
     \end{Verbatim}

     There are multiple problems with this approach, where even the obviously
     huge memory requirement for trying to store $s \in \mathbb{S}$ in a
     tabular structure is not the largest one. An even greater problem is that
     Python's semantics of \texttt{Dictionaries} are not designed to consider
     similarities. Instead, they are designated as key-value pairs. For a Python
     \texttt{dictionary} the two states $s_1 = (1.0, 1.0, 1.0, 1.0)$ and
     $s_2 = (1.0001, 1.0, 1.0, 1.0)$ are completely different and not at all
     similar. In contrast, for a control algorithm that balances a pendulum,
     a cart being at x-position $1.0$ is in a very similar situation (state)
     to a cart being at x-position $1.0001$. So Python's
     \texttt{Dictionaries} cannot be used to model $Q(s, a)$. Also,
     trying to use other tabular methods such as Python's \texttt{Lists},
     creates many other challenges. 

     This is why \emph{linear regression} has been chosen to represent and
     model the function $Q(s, a)$. As we only have two possible
     states $a \in \{0, 1\}$, no general purpose implementation of $Q(s, a)$
     has been done. Instead, two discrete functions
     $Q_{a=0}(s)$ and $Q_{a=1}(s)$ are modeled using a linear regression
     algorithm from \textsc{scikit-learn} called
     \texttt{SGDRegressor}\footnote{SGD stands for
     Stochastic Gradient Descent}, which is capable of performing
     online learning. In online machine learning, data
     becomes available step by step and is used to update the best predictor
     for future data at each step as opposed to batch learning, where the
     best predictor is generated by using the entire training data set
     at once.

     Given $m$ input features $f_i~(i = 1\dots m)$ and 
     corresponding coefficients $\theta_i~(i = 0\dots m)$, linear regression 
     can predict $\hat{y}$ as follows:
     \[
      \hat{y} = \theta_0 + \theta_1 f_1 + \theta_2 f_2 + \dots + \theta_m f_m
     \]
     The whole point of \texttt{SGDRegressor} is to 
     iteratively refine the values for all $\theta_i$, as more and more pairs
     of $[\hat{y}, f_i~(i=1\dots m)]$ are generated during each step of each
     episode of the $Q$-learning algorithm with more and more accuracy due
     to the policy iteration. Older pairs are likely less accurate estimates
     and are therefore discounted.

     A natural choice of features for linear regression would be to set
     $m = 4$ and to use the elements of the state $s \in \mathbb{S}$ as the
     input features $f$ of the linear regression:
     \[
      Q_{a=n}(s) = \theta_0 + \theta_1 x + \theta_2 \dot{x} + 
                   \theta_3 \varphi + \theta_4 \dot{\varphi}
     \]
     Experiments performed during the implementation have shown that this
     choice of features does not produce optimal learning results, as it
     would lead to \emph{underfitting}, i.e. the learned model makes too rough
     predictions for the pendulum to be balanced.
     The next section \ref{sss_ftrans} explains this phenomenon.

     \texttt{SGDRegressor} offers built-in mechanisms to handle the
     learning rate $\alpha$. When constructing the
     \texttt{SGDRegressor} object, $\alpha$ can be directly specified as
     a parameter. Therefore, line \ref{a_qmagic} of algorithm
     \ref{a_q-learning} on page \pageref{a_q-learning} is simplified in the
     Python implementation, as "the $\alpha$ used in the $Q$-learning
     algorithm" and "the $\alpha$ used inside the \texttt{SGDRegressor}"
     are semantically identical: The purpose of both of them is to act
     as the learning rate. Thus, the "the $\alpha$ used in the $Q$-learning
     algorithm" can be omitted (means $\alpha$ can be set to $1$):
     \[
      Q_{\text{SGDReg}}(s, a) \gets r + \gamma * Q(s', a_{\text{best}})
     \]
     It needs to be mentioned, that the \texttt{SGDRegressor} is not just
     overwriting the old value with the new value when the above-mentioned
     formula is executed. Instead, when its online learning function
     \texttt{partial\_fit} is called, it improves the existing predictor
     for $Q(s, a)$ by using the new value $r + \gamma * Q(s', a_{\text{best}})$
     discounted by the learning rate $\alpha$.

     In plain english, calling
     \texttt{partial\_fit} is equivalent to ``the linear regressor that
     is used to represent $Q(s, a)$ in Python updating its knowledge about
     the action value function $Q$ for the state $s$ in which action $a$ has
     been taken by a new estimate without forgetting what has been previously
     learned. The new estimate that is used to update $Q$ consists of the
     short-term reward $r$ that came as a result of taking action $a$
     while being in step $s$ plus the discounted estimated long-term reward
     $\gamma * Q(s', a_{\text{best}})$ that is obtained by acting as if after 
     action $a$ has been taken, in future always the best possible future action
     will be taken.''

     \begin{Verbatim}[numbers=left,frame=single,labelposition=all,
                      label=\scriptsize analog-cartpole.py]
# List of possible actions that the RL agent can perform in
# the environment. For the algorithm, it doesn't matter if 0
# means right and 1 left or vice versa or if there are more
# than two possible actions
env_actions = [0, 1]

[...]

# we use one Linear Regression per possible action to model
# the Value Function for this action, so rbf_net is a list;
# SGDRegressor allows step-by-step regression using
# partial_fit, which is exactly what we need to learn
rbf_net = [SGDRegressor(eta0=ALPHA, 
                        power_t=ALPHA_DECAY,
                        learning_rate='invscaling',
                        max_iter=5, tol=float("-inf"))
            for i in range(len(env_actions))]

[...]

# learn Value Function for action a in state s
def rl_set_Q_s_a(s, a, val):
    rbf_net[a].partial_fit(rl_transform_s(s),
                           rl_transform_val(val))

[...]

# Learn the new value for the Value Function
new_value = r + GAMMA * max_q_s2a2
rl_set_Q_s_a(s, a, new_value)
     \end{Verbatim}

     The source snippet shows that there is a regular Python \texttt{List}
     which contains two\footnote{\texttt{for i in range(len(env\_actions))}
     leads to two iterations as \texttt{env\_actions} contains two
     elements} objects of type \texttt{SGDRegressor}. Therefore
     \texttt{rbf\_net[n]} can be considered as the representation of
     $Q_{a=n}(s)$ in memory. For increasing the accuracy of $Q$ during the
     learning process, the function \texttt{rl\_set\_Q\_s\_a(s,~a,~val)}
     which itself uses \texttt{SGDRegressor}'s \texttt{partial\_fit} function
     is called regularly at each step of each episode.
     Therefore lines 29--30 are equivalent to line \ref{a_qmagic} of
     algorithm \ref{a_q-learning}.
    \subsubsection{Feature transformation to avoid underfitting}
     \label{sss_ftrans}
     Underfitting occurs when a statistical model or machine learning
     algorithm cannot capture the underlying trend of the data. Intuitively,
     underfitting occurs when the model or the algorithm does not fit the
     data well enough, because the complexity of the model is too low.
     As shown in section \ref{s_analog}, the inverse pendulum is a non-linear
     function so one might think that this is the reason that a simple linear
     regression which tries to match the complexity of a non-linear function
     using $s \in \mathbb{S}$ as feature set might be prone to underfitting.

     In general, this is not the case. As shown at the beginning of this
     section on page \pageref{ss_f_theta_s} where a control algorithm is
     introduced, that is based on a randomly found $\vec{\theta}$ and that
     controls the cart via the linear dot multiplication
     $\vec{\theta} \cdot \vec{s}$ and
     where both vectors consist of mere four elements, simple linear functions
     can absolutely control a complex non-linear phenomenon. So linear
     control functions per se are not the reason for underfitting.

     Instead, the reason why underfitting occurs in the context of the
     $Q$-learning algorithm is that the very concept of reinforcement learning's
     value function, which is a general purpose concept for machine learning,
     creates overhead and complexity. And this complexity needs to be matched
     by the statistical model - in our case linear regression - that is
     chosen to represent $Q(s, a)$. The value function is more complex than a
     mere control policy like the one that is described on page
     \pageref{ss_f_theta_s}, because it not only contains the information
     necessary to control the cart but it also contains the information about
     the expected future reward. This surplus of information needs to
     be stored somewhere (i.e.\ needs to be fitted by the model of choice).

     The model complexity of linear regression is equivalent to the number
     $m$ of input features $f_i~(i = 1\dots m)$ in the
     linear regression equation
     $\hat{y} = \theta_0 + \theta_1 f_1 + \theta_2 f_2 + \dots + \theta_m f_m$
     .  

     Finding out the optimal threshold of the model complexity necessary to
     avoid underfitting is a hard task that has not been solved in data
     science in general at the time of writing. Therefore the model complexity
     is another one of the many \emph{hyper parameters} that need to be
     found and fine-tuned before the actual learning process begins. Finding
     the right hyper parameters often requires a combination of intuition
     and trial and error.

     Consequently the challenge that had to be solved by the authors was:
     When the model complexity of linear regression needs to be increased,
     more features are needed. But how can more than four features be
     generated, given that $s \in \mathbb{S}$ only consists of the four
     features $x, \dot{x}, \varphi, \dot{\varphi}$?

     The solution is to perform a \emph{feature transformation}, where the
     number of new features after transforming $s \in \mathbb{S}$ is
     significantly bigger than four:
     \[
      (x, \dot{x}, \varphi, \dot{\varphi})
      \rightarrow (f_1, \dots, f_m)~(i=1\dotsm,~m >> 4)
     \]
     Since the open question here is ``how much means significantly?'',
     one of the requirements for a good feature transformation function 
     is flexibility in the sense that it must be as easy to adjust the
     hyperparamter ``model complexity $m$'' as it is to adjust some constants
     in the Python code (versus finding a completely new feature
     transformation function each time $m$ needs to be increased
     or decreased).

     It is a best practice of feature engineering that a feature
     transformation introduces a certain non-linearity. This can be done using
     many different options such as using a polynomial over the original
     features. 

     \emph{Radial Basis Functions} (RBF) are another option and have been
     chosen as a means of feature transformation that allows to increase
     the model complexity $m$  of the linear regression.
     An RBF is a function that maps a vector to
     a real number by calculating (usually) the Euclidian distance from the
     function's argument to a previously defined \emph{center} which is
     sometimes also called an \emph{exemplar}. This distance is then used
     inside a \emph{kernel} function to obtain the output value of the RBF.

     By using a large number of RBF transformations, where each RBF uses a
     different, randomly chosen, center,
     a high number $m$ of input features $f_i$
     can be generated for the linear regression. In other words: A feature
     map of $s$ is generated by applying $m$ different RBFs with $m$
     different random centers on $s$:
     \begin{align}
      s &= (x, \dot{x}, \varphi, \dot{\varphi}) \nonumber\\   
      center_i &= \text{random}(x_i, \dot{x_i}, \varphi_i, \dot{\varphi_i})~
                \text{for all}~i = 1\dots m, \nonumber\\
      \delta_i &= \text{distance}_{euclid_i} = 
                \left\lVert (s - center_i) \right\rVert \nonumber\\
      RBF_i&:  s \in \mathbb{S} \rightarrow y_i \in \mathbb{R} \nonumber\\
      RBF_i&:  e^{-(\beta \delta_i)^2} \rightarrow y_i \label{f_RBF}
     \end{align}
     $\beta$ is a constant for all $i = 1\dots m$ that defines the shape
     of the bell curve described by the \emph{Gaussian} RBF kernel applied
     here. So in summary, the method described here is generating $m$ features
     from the original four features of $s$ where due to the fact that the
     Eucledian distance to the centers is used, similar states $s$ are
     yielding similar transformation results $y$. As long as this similarity
     premise holds and as long as the feature transformation is not just a
     linear combination of the original features but adds non-linearity,
     it actually does not matter for the purposes of adding more model
     complexity to the linear regression, what kind of feature transformation
     is applied. RBFs are just one example.

     Due to the fact that it is beneficial\footnote{High Python program
     execution performance is helpful to achieve a high sampling rate as
     described in section \ref{sss_states} and therefore a high accuracy}
     in the context of the analog computer simulation
     to achieve near real-time performance of the Python software,
     the above-mentioned RBF transformation has not been implemented
     verbatim in Python. This would have been too slow. Instead, the 
     \textsc{scikit-learn} class \texttt{RBFSampler} has been used, which 
     generates the feature map of an RBF kernel using a Monte Carlo
     approximation of its Fourier transform. The experiments have shown that
     \texttt{RBFSampler} is fast enough and that the approximation is
     good enough\footnote{See also
     \protect\url{https://tinyurl.com/RBFSampler}}.

     \begin{Verbatim}[numbers=left,frame=single,labelposition=all,
                      label=\scriptsize analog-cartpole.py]
# The following four constants are tunable hyperparameters.
# Please note that the source code uses the term GAMMA for
# denoting what we are calling BETA in this chapter: The
# shape of the bell curve.
RBF_EXEMPLARS   = 250  # amount of exemplars per "gamma 
                       # instance" of the RBF network
RBF_GAMMA_COUNT = 10   # amount of "gamma instances", i.e.
                       # RBF_EXEMPLARS*RBF_GAMMA_COUNT feat.
RBF_GAMMA_MIN   = 0.05 # minimum gamma, linear interpolation
                       # between min and max
RBF_GAMMA_MAX   = 4.0  # maximum gamma

[...]

# create scaler and fit it to the sampled observation space
scaler = StandardScaler() 
scaler.fit(clbr_res)

[...]

# the RBF network is built like this: create as many
# RBFSamplers as RBF_GAMMA_COUNT and do so by setting the
# "width" parameter GAMMA of the RBFs as a linear
# interpolation between RBF_GAMMA_MIN and RBF_GAMMA_MAX
gammas = np.linspace(RBF_GAMMA_MIN,
                     RBF_GAMMA_MAX,
                     RBF_GAMMA_COUNT)
models = [RBFSampler(n_components=RBF_EXEMPLARS,
                     gamma=g) for g in gammas]

# we will put all these RBFSamplers into a FeatureUnion, so
# that our Linear Regression can regard them as one single
# feature space spanning over all "Gammas"
transformer_list = []
for model in models:
    # RBFSampler just needs the dimensionality,
    # not the data itself
    model.fit([[1.0, 1.0, 1.0, 1.0]]) 
    transformer_list.append((str(model), model))
    # union of all RBF exemplar's output
    rbfs = FeatureUnion(transformer_list)     

[...]

# transform the 4 features (Cart Position, Cart Velocity,
# Pole Angle and Pole Velocity At Tip) into
# RBF_EXEMPLARS*RBF_GAMMA_COUNT distances from the 
# RBF centers ("Exemplars")
def rl_transform_s(s):
    # during calibration, we do not have a scaler, yet
    if scaler == None:
        return rbfs.transform(
               np.array(s).reshape(1, -1))
    else:
        return rbfs.transform(
               scaler.transform(
               np.array(s).reshape(1, -1)))

# SGDRegressor expects a vector, so we need to transform
# our action, which is 0 or 1 into a vector
def rl_transform_val(val):
    return np.array([val]).ravel()
     \end{Verbatim}

     Section \ref{sss_Q} is missing an explanation for the two functions
     \texttt{rl\_transform\_s} and \texttt{rl\_transform\_val} used in
     lines 23 and 24 of the source code snippet shown there.
     The latter one is just a technical necessity,
     because \texttt{SGDRegressor} expects a vector. In our model, the action
     is not a vector but an integer, so \texttt{rl\_transform\_val}
     ensures, that we transform the integer 
     $a$ into the vector $\vec{a} = \left( a \right)$.

     The actual feature transformation is taking place in
     \texttt{rl\_transform\_s}. As described above,
     the \textsc{scikit-learn} class \texttt{RBFSampler} is used
     for performance reasons.
     The model complexety $m$ is defined by two constants in Python:
     \texttt{RBF\_EXEMPLARS} and \texttt{RBF\_GAMMA\_COUNT}\footnote{The
     source code calls $\beta$ not \texttt{BETA} in this context 
     but \texttt{GAMMA}. In contrast, $\gamma$ is used to denote the
     discount factor in this chapter.} and $m$ is the product of both of them.
     As an enhancement to the formula
     $RBF_i:  e^{-(\beta \delta_i)^2} \rightarrow y_i$ explained
     on page \pageref{f_RBF}, where the shape parameter $\beta$ of the
     Gaussian curve is kept constant, the solution here creates a linear
     interpolation between the value \texttt{RBF\_GAMMA\_MIN} and
     \texttt{RBF\_GAMMA\_MAX} consisting of \texttt{RBF\_GAMMA\_COUNT}
     distinct values for $\beta$. The overall result of $m$ RBF
     transformations is united in a Python \texttt{FeatureUnion} so that
     the function \texttt{rl\_transform\_s} can conveniently call the
     object \texttt{rbfs} to execute the transformation.

     The experiments have shown, that the learning efficiency is higher,
     when the values that are coming from the analog computer are processed
     as they are instead of being scaled before.
     Therefore, the full source code on GitHub (see footnote \ref{fn_github})
     contains a boolean switch called \texttt{PERFORM\_CALIBRATION} that is
     set to \texttt{False} in the final version. This means, that
     the \texttt{if} branch shown in lines 48 and 49 is the one performing the
     feature transformation.
    \subsubsection{Decaying $\alpha$ and $\varepsilon$ to improve learning
     and to avoid overfitting}
     Our experiments have shown that it makes sense to start with relatively
     high values for the learning rate $\alpha$ and the explore vs. exploit
     probability $\varepsilon$ and then to decay them over time. This is the
     equivalent of trying many different options at the beginning when there
     is still very little knowledge available (high $\varepsilon$ means to
     favor explore over exploit) and learning quickly
     from those explorations (high $\alpha$). Later when knowledge has
     accumulated, it takes more effort to modify facts that have been
     learned earlier (low $\alpha$).

     SGDRegressor has a built in function to decay $alpha$ over time. The
     $\varepsilon$ decay has been implemented manually. The following code
     snippet shows the actual parameters used in the implementation.

     \begin{Verbatim}[numbers=left,frame=single,labelposition=all,
                      label=\scriptsize analog-cartpole.py]
GAMMA            = 0.999 # discount factor for Q-learning
ALPHA            = 0.6   # initial learning rate
ALPHA_DECAY      = 0.1   # learning rate decay
EPSILON          = 0.5   # randomness for epsilon-greedy
EPSILON_DECAY_t  = 0.1   # decay parameter for epsilon
EPSILON_DECAY_m  = 10    # ditto
     \end{Verbatim}

     Decaying $\alpha$ and $\varepsilon$ is not only used to improve the
     learning itself, but also to avoid \emph{overfitting} the model.
     Overfitting is described\footnote{\protect\url{
     https://www.lexico.com/definition/overfitting
     }} as the ``production of an analysis which
     corresponds too closely or exactly to a particular set of data, and may
     therefore fail to fit additional data or predict future observations
     reliably''.

     Another mechanism that was used to avoid overfitting is to make sure
     that the system is ``not learning for too long''. The concrete meaning of
     ``too long'' is - as many things in machine learning - another
     hyper parameter. For being able to persistently access the (near to)
     optimal learning state, the Python implementation supports
     a command line parameter, that forces the current state of the ``brain''
     to be saved every \texttt{PROBE} episodes, whereas \texttt{PROBE} is
     a constant (hyper parameter).
  \subsection{Hybrid Interface} \label{ss_hybrid}
   The \textsc{Analog Paradigm Model-1}\footnote{\protect\url{
   http://analogparadigm.com/products.html}} analog computer used by
   the authors incorporates a \emph{Hybrid Controller} for connecting
   the \textsc{Model-1} to digital computers. It uses an
   \emph{RS232 over USB} mechanism that is compatible with most PC operating
   systems, in the sense that the PC operating system is able to provide a
   virtual serial port that behaves exactly as if the analog and the digital
   computer where connected by an RS232 cable instead of an USB cable.
   This is why, in Python, the communication uses the
   \texttt{pySerial}\footnote{\protect\url{
   https://pypi.org/project/pyserial}} library. The following code snippet
   shows the setup parameters. Note that the Hybrid Controller is
   communicating at 250,000 baud.

   \begin{Verbatim}[numbers=left,frame=single,labelposition=all,
                    label=\scriptsize analog-cartpole.py]
# Hybrid Controller serial setup
HC_PORT             = "/dev/cu.usbserial-DN050L1O"
HC_BAUD             = 250000        
HC_BYTE             = 8
HC_PARITY           = serial.PARITY_NONE
HC_STOP             = serial.STOPBITS_ONE
HC_RTSCTS           = False
HC_TIMEOUT          = 2

[...]

hc_ser = serial.Serial( port=HC_PORT,
                        baudrate=HC_BAUD,
                        bytesize=HC_BYTE,
                        parity=HC_PARITY,
                        stopbits=HC_STOP,
                        rtscts=HC_RTSCTS,
                        dsrdtr=False,
                        timeout=HC_TIMEOUT)
   \end{Verbatim}

   Analog computers operate in different modes. In this machine
   learning implementation, the Python script controls the analog computer's
   mode of operation according to the needs of the algorithm:
   \begin{description}
    \item [Initial Condition] marks the beginning of an episode.
     The pendulum is in an upright position.
    \item [Operate] is the standard mode of operation, where the analog
     computer is running the simulation in real-time.
    \item [Halt] means that the simulation is paused and that it can be
     resumed any time by returning to the \emph{Operate} mode.
   \end{description}
   The Hybrid Interface accepts certain mode-change commands via the serial
   line to put the analog computer into the appropriate mode. Moreover,
   there are several commands to read data from the various
   \emph{computing elements} in the analog computer.
   All the computing elements can be referenced by unique addresses.
   For being able to influence the calculations of the analog computer,
   the Hybrid Controller provides analog and digital inputs and outputs.
   For the purposes of this problem, only two digital outputs are needed
   to drive the model:\footnote{See section \ref{s_analog}.}
   \begin{description}
    \item [Digital Out \#0] is used to set the direction from which the
     force is being applied when pushing the cart
    \item [Digital Out \#1] makes sure that a force is applied to the cart as
     long as it is being set to \texttt{1}
   \end{description}

   \begin{Verbatim}[numbers=left,frame=single,labelposition=all,
                    label=\scriptsize analog-cartpole.py]
# Addresses of the environment/simulation data
HC_SIM_X_POS        = "0223" # address cart x
HC_SIM_X_VEL        = "0222" # address cart x-velocity
HC_SIM_ANGLE        = "0161" # address pendulum angle
HC_SIM_ANGLE_VEL    = "0160" # address pend. angular vel.

HC_SIM_DIRECTION_1  = "D0"   # dout: cart direct. = 1
HC_SIM_DIRECTION_0  = "d0"   # dout: cart direct. = 0
HC_SIM_IMPULSE_1    = "D1"   # dout: apply force
HC_SIM_IMPULSE_0    = "d1"   # dout: apply NO force

# Model-1 Hybrid Controller: commands
HC_CMD_RESET        = "x"    # reset hybrid controller
HC_CMD_INIT         = "i"    # initial condition
HC_CMD_OP           = "o"    # start to operate
HC_CMD_HALT         = "h"    # halt/pause
HC_CMD_GETVAL       = "g"    # set address of analog
                             # computing element and
                             # return value and ID
HC_CMD_BULK_DEFINE  = "G"    # set addresses of multiple
                             # elements to be returned
                             # in a bulk transfer via "f"
HC_CMD_BULK_FETCH   = "f"    # fetch values of all
                             # addresses defined by "G"
[...]

def hc_send(cmd):
    hc_ser.write(cmd.encode("ASCII"))

def hc_receive():
    # HC ends each communication with "\n",
    #so we can conveniently use readline
    return hc_ser.readline().
                  decode("ASCII").
                  split("\n")[0]
   \end{Verbatim}

   As described in section \ref{sss_states}, the digital computer samples
   the analog computer's continuous simulation by repeatedly reading the
   simulation state variables, to generate the discretization
   needed for $Q$-learning.
   The Hybrid Controller supports this effort by providing a bulk readout
   function that returns the simulation's overall state with as little
   overhead as possible. This significantly reduces latency during the learn
   and control loop, and thus - as experiments have shown - improves the
   efficiency and convergence speed of the $Q$-learning algorithm.

   The bulk mode is activated by sending a bulk definition command that
   defines a \emph{readout group} on the Hybrid Controller. After this has
   been done, the bulk read is triggered by sending a short
   (single character) fetch command. The following code snippet illustrates
   the concept and the process.

   \begin{Verbatim}[numbers=left,frame=single,labelposition=all,
                    label=\scriptsize analog-cartpole.py]
#define a readout group in the Hybrid Controller
hc_send(HC_CMD_BULK_DEFINE + HC_SIM_X_POS + ';'
                           + HC_SIM_X_VEL + ';'
                           + HC_SIM_ANGLE + ';'
                           + HC_SIM_ANGLE_VEL + '.')

# when using HC_CMD_GETVAL,
#HC returns "<value><space><id/type>\n"
# we ignore <type> but we expect a well formed response
def hc_res2float(str):
    f = 0
    try:
        f = float(str.split(" ")[0])
        return f
    except:
        [...]

# query the current state of the simulation, which consists
# of the x-pos and the the x-velocity of the cart, the angle
# and angle velocity of the pole/pendulum
def hc_get_sim_state():
    # bulk transfer: ask for all values that consitute the
    # state in a bulk using a single fetch command
    if HC_BULK:
        hc_send(HC_CMD_BULK_FETCH)
        (res_x_pos, res_x_vel,
         res_angle, res_angle_vel) = hc_receive().split(';')
        return (hc_res2float(res_x_pos), 
                hc_res2float(res_x_vel),
                hc_res2float(res_angle),
                hc_res2float(res_angle_vel))
    else:
        [...]
   \end{Verbatim}

   The only way the environment in this example can be influenced by
   the reinforcement learning agent is via the predefined actions. In this
   case: ``Push the cart to the left or push it to the right.'' When looking
   at this situation from an analog computer's viewpoint, this is a
   continuous operation: ``To push'' means that a force $F$ that does not
   necessarily need to be constant over time is applied for a certain period
   of time. As described in section \ref{ss_actions} the implementation used
   here is heavily simplified: Two possible actions, one fixed and
   constant force $F$ and a constant period of time where the force is applied
   form the two possible impulses. As the magnitude $\left|F\right|$
   of the force is configured directly at the analog computer, the Python
   code can focus on triggering the impulse using the two above-mentioned
   digital outputs.

   \begin{Verbatim}[numbers=left,frame=single,labelposition=all,
                    label=\scriptsize analog-cartpole.py]
# duration [ms] of the impulse, that influences the cart
HC_IMPULSE_DURATION = 20

[...]                    

# influence simulation by using an impulse to
# push the cart to the left or to the right;
# it does not matter if "1" means left or right
# as long as "0" means the opposite of "1"
def hc_influence_sim(a, is_learning):
    [...]

    if (a == 1):
        hc_send(HC_SIM_DIRECTION_1)
    else:
        hc_send(HC_SIM_DIRECTION_0)
   
    hc_send(HC_SIM_IMPULSE_1)
    sleep(HC_IMPULSE_DURATION / 1000.0)
    hc_send(HC_SIM_IMPULSE_0)

    [...]
   \end{Verbatim}

   In summary it can be stated, that working with an analog computer like
   the \textsc{Model-1} with its Hybrid Controller is quite straightforward.
   The serial communication can be encapsulated in some functions or objects
   and from that moment on, the only thing that one needs to keep in mind is
   the completely asynchronous and parallel nature of the setup. No
   assumptions on certain sequence properties can be made and all the typical
   challenges of asynchronous parallel setups like jitter, race conditions,
   latency, etc.\ can occur and need to be managed properly.
 \section{Results}
  Analog computers outperform digital computers with respect to raw 
  computational power as well as with respect to power efficiency for certain
  problems, such as the simulation of dynamic systems that can be readily 
  described by systems of coupled differential equations. This is due to the
  fact that analog computers are inherently parallel in their operation as they
  do not rely on some sort of an algorithm being executed in a step-wise 
  fashion. Other advantages are their use of a continuous value representation
  and that fact that integration is an intrinsic function.

  Typical analog computer setups tend to be extremely stable and are not prone
  to problems like numerical instabilities etc. Although the precision of an
  analog computer is quite limited compared with a stored-program digital 
  computer employing single or even double precision floating point numbers
  (a precision analog computer is capable of value representation with about 4 
  decimal places), solutions obtained by means of an analog computer will 
  always turn out to be realistic, something that cannot be said of numerical
  simulations where the selection of a suitable integration scheme can have a
  big effect on the results obtained.

  In reinforcement learning scenarios, the actual simulation of the environment
  as shown in figure \ref{pic_rl_basics} on page \pageref{pic_rl_basics}
  is one of the hardest and most time consuming things to do. A simulation 
  implemented on an analog computer behaves much more realistically than one
  performed on a digital computer. It yields unavoidable measurement errors
  as would be present in a real-world-scenario, it is immune to numerical 
  problems etc.
  
  A plethora of RL algorithms such as (Deep) $Q$-learning are
  available and well understood, as for example shown in \cite{sutton}.
  Many Open Source implementations are available for free use. So when it
  comes to practially apply RL to solve challenges in rapidly advancing
  fields such as autonomous cars, robot navigation, coupling of human nervous
  signals to artificial limbs and the creation of personalized medical
  treatments via protein folding, a new project
  very often starts with the question ``how can we simulate the environment
  for our agent to learn from it?'' 

  The authors have shown in this chapter that this challenge can be very
  successfully tackled using analog computers, so that a next generation of
  analog computers based on VLSI analog chips could help to overcome 
  this mainstream challenge of reinforcement learning by propelling the
  speed and accuracy of environment simulation for RL agents to a new
  level. This would result in faster development cycles for RL enabled
  products and therefore could be one of many catalysts in transforming
  reinforcement learning from a research discipline to an economically viable
  building block for the truly intelligent device of the future.
 
\end{document}